\documentclass[10pt,twocolumn,letterpaper]{article}

\usepackage{cvpr}      

\usepackage{graphicx}
\usepackage{amsmath}
\usepackage{amssymb}
\usepackage{booktabs}
\usepackage[export]{adjustbox}
\usepackage{enumitem}
\usepackage[table]{xcolor}
\usepackage{multirow}
\usepackage[accsupp]{axessibility}

\makeatletter
\newcommand{\thickhline}{%
	\noalign {\ifnum 0=`}\fi \hrule height 1pt
	\futurelet \reserved@a \@xhline
}
\makeatother
\newcommand{\tablestyle}[2]{\setlength{\tabcolsep}{#1}\renewcommand{\arraystretch}{#2}\centering\footnotesize}

\makeatletter
\newcommand{\printfnsymbol}[1]{%
  \textsuperscript{\@fnsymbol{#1}}%
}
\makeatother

\definecolor{mygray}{gray}{.9}
\definecolor{sota_blue}{HTML}{0071bc}

\definecolor{citecolor}{RGB}{34,139,34}
\usepackage[pagebackref,breaklinks,colorlinks,citecolor=citecolor]{hyperref}

\usepackage[capitalize]{cleveref}
\crefname{section}{Sec.}{Secs.}
\Crefname{section}{Section}{Sections}
\Crefname{table}{Table}{Tables}
\crefname{table}{Tab.}{Tabs.}

\def\cvprPaperID{600} 
\def\confName{CVPR}
\def\confYear{2022}

\begin{document}

\title{Language-Bridged Spatial-Temporal Interaction for \\Referring Video Object Segmentation}

\author{Zihan Ding\textsuperscript{\rm 1,4,5} \quad Tianrui Hui\textsuperscript{\rm 2,3} \quad Junshi Huang\textsuperscript{\rm 4} \quad Xiaoming Wei\textsuperscript{\rm 4} \quad Jizhong Han\textsuperscript{\rm 2,3} \quad Si Liu\textsuperscript{\rm 1,5}\thanks{Corresponding author} \\
\textsuperscript{\rm 1} Institute of Artificial Intelligence, Beihang University\\
\textsuperscript{\rm 2} Institute of Information Engineering, Chinese Academy of Sciences\\
\textsuperscript{\rm 3} School of Cyber Security, University of Chinese Academy of Sciences\\
\textsuperscript{\rm 4} Meituan \quad \textsuperscript{\rm 5} Hangzhou Innovation Institute, Beihang University\\
}

\maketitle

\begin{abstract}

Referring video object segmentation aims to predict foreground labels for objects referred by natural language expressions in videos.
Previous methods either depend on 3D ConvNets or incorporate additional 2D ConvNets as encoders to extract mixed spatial-temporal features.
However, these methods suffer from spatial misalignment or false distractors due to delayed and implicit spatial-temporal interaction occurring in the decoding phase.
To tackle these limitations, we propose a Language-Bridged Duplex Transfer (LBDT) module which utilizes language as an intermediary bridge to accomplish explicit and adaptive spatial-temporal interaction earlier in the encoding phase.
Concretely, cross-modal attention is performed among the temporal encoder, referring words and the spatial encoder to aggregate and transfer language-relevant motion and appearance information.
In addition, we also propose a Bilateral Channel Activation (BCA) module in the decoding phase for further denoising and highlighting the spatial-temporal consistent features via channel-wise activation.
Extensive experiments show our method achieves new state-of-the-art performances on four popular benchmarks with 6.8\% and 6.9\% absolute AP gains on A2D Sentences and J-HMDB Sentences respectively, while consuming around 7$\times$ less computational overhead~\footnote{\url{https://github.com/dzh19990407/LBDT}}. 

\end{abstract}

\section{Introduction}
\label{sec:intro}

Referring video object segmentation (RVOS), which aims to segment the target object referred by a natural language expression in video frames, is an emerging task at the intersection of computer vision and natural language processing.
Different from semi-automatic video object segmentation (SVOS)~\cite{wang2021swiftnet,park2021learning,duke2021sstvos,ge2021video}, where the target object is referred by the manually annotated mask in the first frame, RVOS is more challenging for identifying the targets due to the variance of free-form expressions. Providing a more natural way for human-computer interaction, RVOS opens up a wide range of applications including language-based video editing~\cite{fu2021language}, language-guided video summarization~\cite{narasimhan2021clip}, and video question answering~\cite{jang2017tgif,xu2016msr},~\etc.

\begin{figure}[t]
	\centering
		\includegraphics[width=\linewidth]{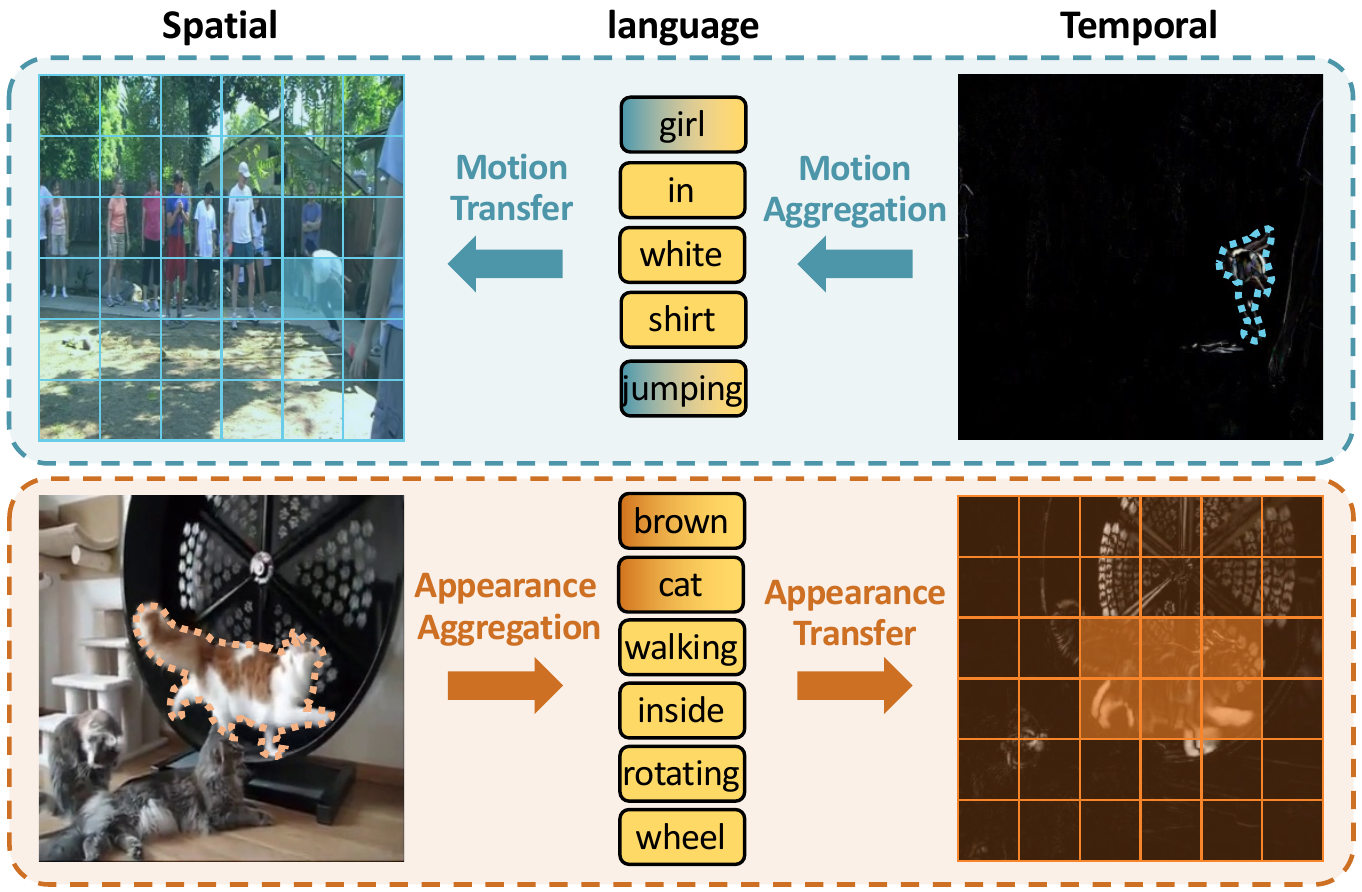}
	\caption{Illustration of our main idea. (Top) For \textit{temporal}$\rightarrow$\textit{language}$\rightarrow$\textit{spatial} transfer, the referring words (\eg, ``jumping'') can aggregate language-relevant motion information from the temporal features, which can help the spatial encoder recognize correct actions. (Bottom) For \textit{spatial}$\rightarrow$\textit{language}$\rightarrow$\textit{temporal} transfer, the referring words (\eg, ``brown'') can aggregate the language-relevant appearance information from the spatial features, which helps the temporal encoder remove the disturbance of background motion (\eg, the rotating wheel).}

	\label{fig:motivation}
	
\end{figure}

The keys to solving RVOS are \textit{spatial-temporal interaction} and \textit{cross-modal alignment}~\cite{HuiH0DLWH021,ye2021referring}. Existing methods mainly focus on the latter and design several mechanisms (\eg, cross-modal attention~\cite{WangDYT19,NingXW020}, capsule routing~\cite{mcintosh2020visual}, and dynamic convolution~\cite{GavrilyukGLS18, WangDMY20}) to mine the semantic correspondence between vision and language modalities. However, all these methods have limitations on spatial-temporal 
interaction due to the reliance on 3D ConvNets (\eg, I3D~\cite{CarreiraZ17}). Concretely, since the poses and locations of moving objects vary in adjacent frames, aggregating spatially misaligned multi-frame features via 3D operators (\eg, 3D convolution and 3D pooling) may confuse the original appearance information in the target frame, leading to inaccurate segmentation results.

To alleviate this phenomenon, CSTM~\cite{HuiH0DLWH021} introduces an additional 2D spatial encoder (\eg, ResNet~\cite{He2016CVPR}) to extract undisturbed appearance information of the target frame, which is fused with features of the temporal encoder in the later decoding phase.
However, the spatial encoder of CSTM lacks motion information since it doesn't explicitly interact with the temporal encoder, making it hard to distinguish among objects with similar appearances while performing different actions.
Thus, it tends to generate high responses on false objects and introduces noises inevitably.

In this paper, we argue that an explicit interaction between spatial and temporal features should be established earlier in the encoding phase, forming a more sufficient and effective information exchange process between encoders. 
Moreover, naive spatial-temporal interaction still tends to introduce noises due to redundant information contained in language-irrelevant distractors. 
Therefore, we believe \textit{the language expression can be exploited as the medium to bridge spatial and temporal interaction}, where only language-relevant information can be transferred between encoders for valid context aggregation.
To this end, we propose a novel Language-Bridged Duplex Transfer (LBDT) module for effective spatial-temporal interaction in the encoding phase.
As illustrated in Figure~\ref{fig:motivation}, motion information from the temporal encoder is first aggregated to the referring words by cross-modal attention.
Then, the spatial encoder can obtain language-relevant motion clues from the referring words by reversed cross-modal attention, which assists in identifying the referred object by recognizing correct actions (Figure~\ref{fig:motivation} top).
Similarly, appearance information from the spatial encoder is also transferred to the temporal encoder through the language bridge, which facilitates the temporal encoder to distinguish language-relevant foreground objects from complex backgrounds (Figure~\ref{fig:motivation} bottom).
In addition, we also remove the dependence on 3D ConvNets and approximate motion information with frame difference processed by a 2D ConvNet.
By this means, the model complexity is significantly reduced as 2D ConvNet occupies nearly 30$\times$ less computational overhead compared to 3D ConvNet (e.g., 3.6 \vs 107.9 GFLOPs)~\cite{chen2018multi}.

To exploit rich multi-scale contexts of hierarchical visual features for finer mask predictions, we also propose a Bilateral Channel Activation (BCA) module to adjust different feature channels in the decoding phase. Concretely, we first upsample and add multi-level features together in temporal and spatial decoders respectively to obtain the decoded features, on which linguistic features are utilized to filter out language-irrelevant motion and appearance information by channel-wise activation. Meanwhile, the global contexts of the decoded features are further extracted to activate the spatial-temporal consistent channels for highlighting features of the referred object.

In a nutshell, our contributions are three-fold:
1) We propose a Language-Bridged Duplex Transfer (LBDT) module to conduct spatial-temporal interaction explicitly between two independent 2D ConvNets in the encoding phase for RVOS, where we use referring words as the medium to transfer language-relevant motion and appearance information.
2) In the decoding phase, we propose a Bilateral Channel Activation (BCA) module to obtain the language-denoised spatial-temporal consistent features for segmenting the referred object.
3) Extensive experiments show that our proposed method outperforms previous methods on four popular RVOS benchmarks, with significant AP gains of 6.8\% on A2D Sentences and 6.9\% on J-HMDB Sentences, while consuming around 7$\times$ less computational overhead.

\section{Related Work}
\label{sec:related}

\subsection{Referring Image Segmentation}
\label{sec:related:ris}

The goal of referring image segmentation (RIS) is to segment the corresponding object referred by a natural language expression in a static image. The task is first proposed by Hu~\etal~\cite{hu2016segmentation}, where visual features extracted by FCN~\cite{long2015fully} and language features extracted by LSTM~\cite{hochreiter1997long} are directly concatenated and fused to form the cross-modal features, based on which the segmentation mask of the referred object is predicted.
Most recent methods follow this one-stage paradigm and design sophisticated ways of cross-modal interaction involving fine-grained dependency modeling and structural analysis~\cite{li2018referring,ye2021referring,jing2021locate,huang2020referring,hui2020linguistic}. Moreover, as RIS and referring expression comprehension (\ie, predicting the bounding box for the referred object instead of mask) are highly related, MCN~\cite{luo2020multi} proposes a multi-task collaborative network to achieve joint learning of the two tasks.
In this paper, we also follow the one-stage paradigm but focus more on realizing effective and efficient spatial-temporal interaction under the mediation of language.

\subsection{Referring Video Object Segmentation}
\label{sec:related:rvos}

\begin{figure*}[t]
	\centering
		\includegraphics[width=0.85\linewidth]{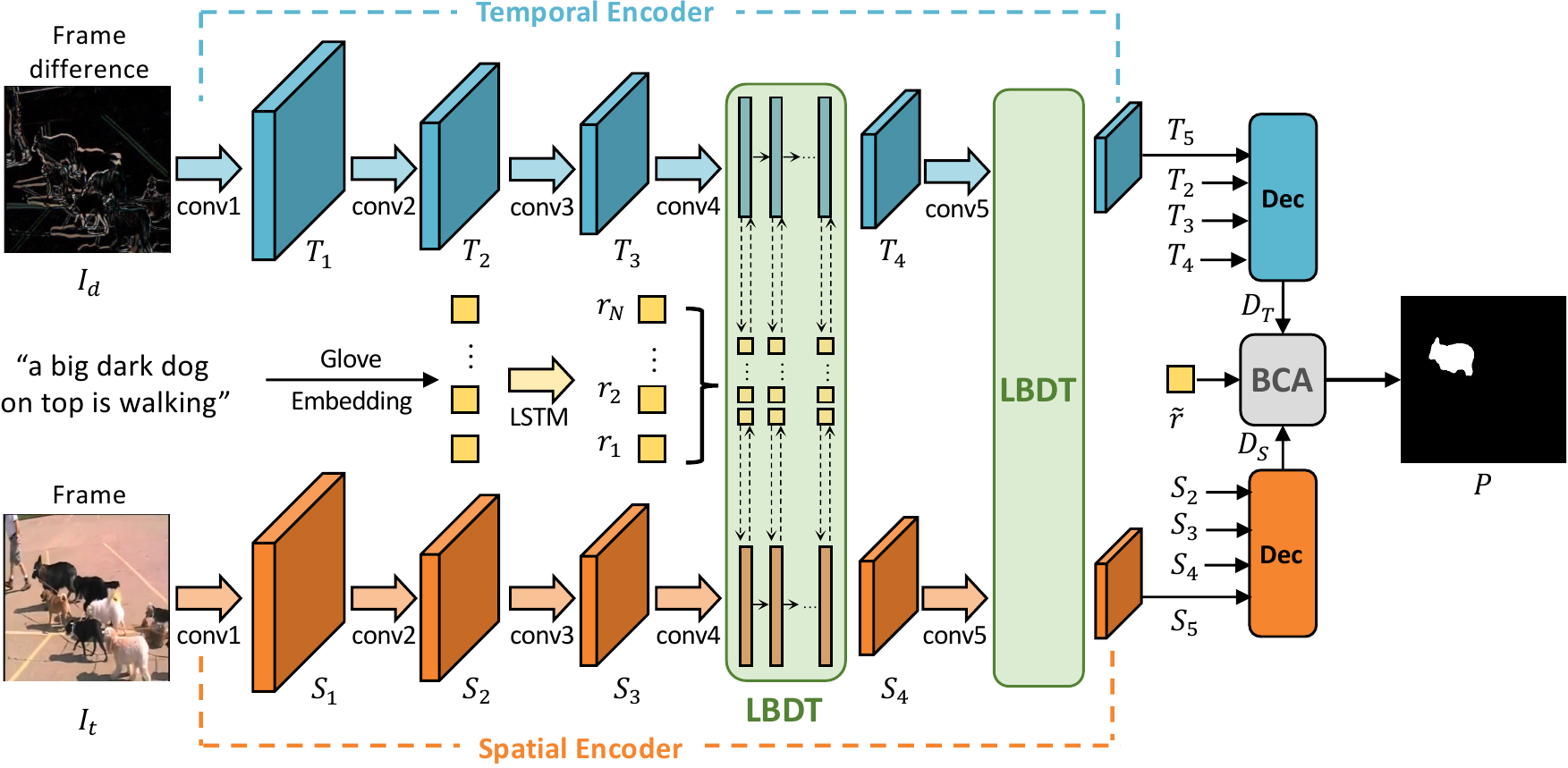}
	\caption{Overview of our proposed method. We feed the target frame $I_t$ and its frame difference $I_d$ into the spatial encoder (bottom) and temporal encoder (top) respectively. And in the LBDT module, we stack several LBDT layers to conduct spatial-temporal interaction with referring words as the medium. In the decoding phase, we denoise the language-irrelevant motion and appearance information and activate the spatial-temporal consistent channels for the decoded spatial features $D_S$ and temporal features $D_T$ respectively in the proposed BCA module. Finally, we apply convolutions and sigmoid function on the outputs of BCA module to get the prediction $P$.}

	\label{fig:framework}
	
\end{figure*}

Referring video object segmentation (RVOS) can be regarded as an extension of RIS, where both motion and appearance information is required to segment the correct object in a dynamic video.
With the availability of diverse benchmarks~\cite{GavrilyukGLS18,khoreva2018video,seo2020urvos}, RVOS has achieved notable progress recently.
Most existing methods mainly feed a video clip centered on the target frame to a 3D ConvNet (\eg, I3D~\cite{CarreiraZ17}), and then obtain mixed spatial-temporal features of the target frame via 3D convolution and pooling.
Similar to RIS methods, they focus on designing different mechanisms for better mining the semantic correspondence between video and language features~\cite{WangDYT19,ye2021referring,NingXW020,GavrilyukGLS18, WangDMY20,mcintosh2020visual}, while neglecting the spatial misalignment issue caused by 3D operators.
Accordingly, CSTM~\cite{HuiH0DLWH021} leverages an additional 2D spatial encoder to complement undisturbed spatial information of the target frame with temporal features in the decoding phase, but still introduce noises with the absence of motion clues in the encoding phase.
In this paper, we propose to establish an explicit spatial-temporal interaction earlier in the encoding phase, where language is exploited as a bridge between spatial and temporal encoders to transfer only language-relevant motion and appearance information while suppressing other irrelevant distractors.

\subsection{Spatial-Temporal Interaction}
\label{sec:related:stm}

Recently, the improvement of spatial-temporal interaction has been widely witnessed in the field of unsupervised video object segmentation~\cite{zhou2020motion,ren2021reciprocal,fragkiadaki2015learning,li2018unsupervised,li2019motion,yan2019semi}.
For example, Zhou~\etal~\cite{zhou2020motion} propose a motion-attentive transition to reinforce spatial-temporal object representations with motion information.
Ren~\etal~\cite{ren2021reciprocal} propose a reciprocal transformation network to discover primary objects by correlating both motion and appearance clues. However, these methods only consider spatial-temporal interaction but ignore the vision-language alignment, while the latter is also crucial for RVOS.
In this paper, we exploit language as a medium to bridge spatial-temporal interaction for extracting comprehensive multimodal representations. 

\section{Method}
\label{sec:method}

The overall architecture of our model is shown in Figure~\ref{fig:framework}.
For the input video clip, we feed its target frame $I_t$ annotated with ground-truth mask and the computed frame difference $I_d$ into two independent ResNet-50~\cite{He2016CVPR} backbones respectively, which are denoted as spatial encoder and temporal encoder in the following.
For the input referring expression, we extract language features from the pretrained GloVe embeddings~\cite{pennington2014glove} with LSTM~\cite{hochreiter1997long}, which are denoted as $R=\{r_n\}_{n=1}^{N}$ where $N$ is the length of referring expression. To explicitly transfer the language-relevant motion and appearance information between the two encoders, we propose a LBDT module and insert it into different encoder stages. In the decoding phase, we integrate multi-scale contexts and propose a BCA module to denoise the language-irrelevant information and activate the spatial-temporal consistent features via channel-wise activation.

\begin{figure*}[t]
	\centering
		\includegraphics[width=0.85\linewidth]{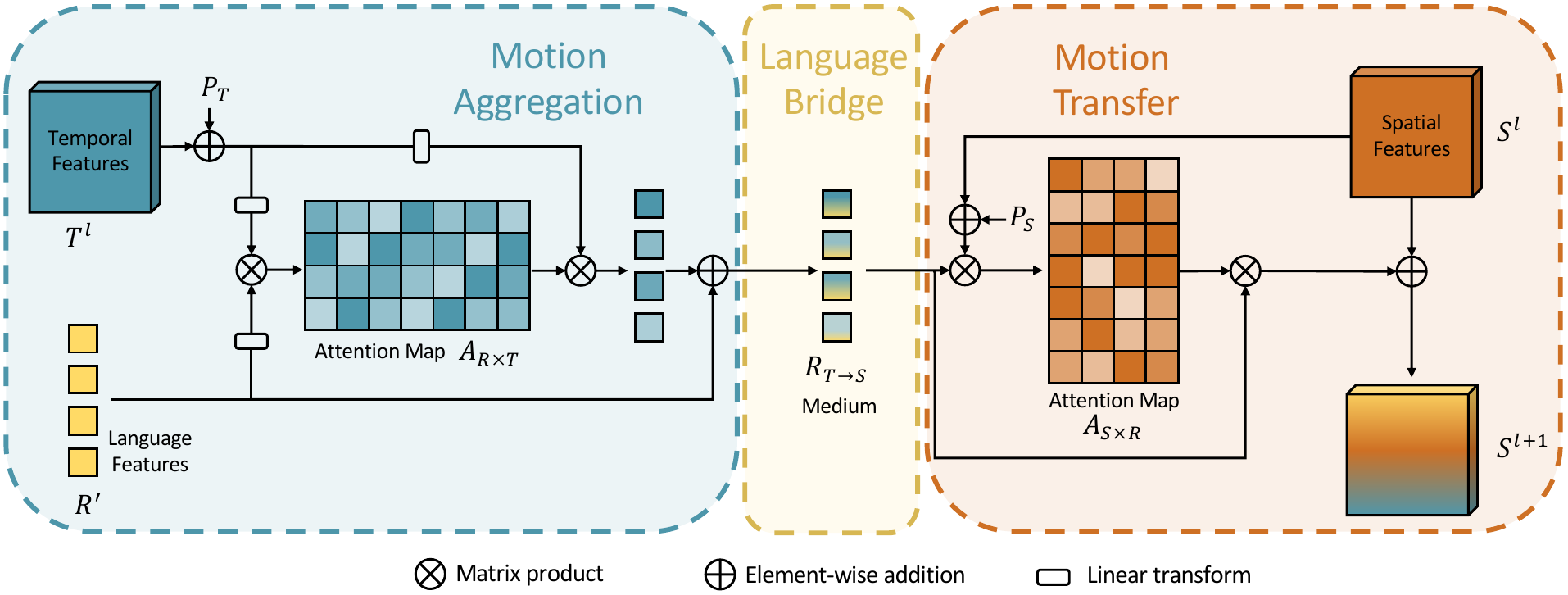}
	\caption{Illustration of the \textit{temporal}$\rightarrow$\textit{language}$\rightarrow$\textit{spatial} information transfer process in our proposed LBDT module. The language-relevant motion information from the temporal features $T^l$ is aggregated into the language medium $R_{T\rightarrow S}$, from which each pixel in the spatial features $S^l$ can select the cross-modal motion information according to the semantic relevance. The \textit{spatial}$\rightarrow$\textit{language}$\rightarrow$\textit{temporal} information transfer process is conducted similarly.}
	\label{fig:lbdt}
	
\end{figure*}

\subsection{Visual and Linguistic Feature Extraction}
\label{sec:method:enc}

Given a video clip, we feed the target frame $I_t\in \mathbb{R}^{3\times H_0\times W_0}$ and the frame difference $I_d=|I_t-I_{t-\delta}|\in \mathbb{R}^{3\times H_0\times W_0}$ to the spatial encoder and temporal encoder respectively, where $\delta$ is the interval between the target frame and the previous frame for calculating the frame difference. Instead of using I3D~\cite{CarreiraZ17} as the temporal encoder, we build our spatial and temporal encoders upon the 2D ResNet-50~\cite{He2016CVPR}. We denote features of the five stages as $\{S_s\}_{s=1}^5, S_s\in \mathbb{R}^{C_s\times H_s\times W_s}$ and $\{T_s\}_{s=1}^5, T_s\in \mathbb{R}^{C_s\times H_s\times W_s}$ for spatial and temporal encoders respectively, where $H_s, W_s=\frac{H_0}{2^s},\frac{W_0}{2^s}$ and $C_s$ are the height, width, and channel numbers of features in the $s$-th stage.

For the referring expression, we embed each word as a $300$-dimensional vector~\cite{pennington2014glove} and use LSTM~\cite{hochreiter1997long} as the text encoder to extract the words features $R=\{r_n\}_{n=1}^N\in \mathbb{R}^{N\times C_m}$, where $N$ is the max length of the referring expressions and $C_m$ is the channel number.

\subsection{Language-Bridged Duplex Transfer}
\label{sec:method:lmdtm}

Our LBDT module aims to explicitly transfer the language-relevant motion and appearance information between temporal and spatial encoders with language as the bridge, where we stack $L$ layers of LBDT module to conduct this duplex transfer approach (Figure~\ref{fig:lbdt}).
To clearly elaborate the transfer process in the LBDT module, we take the $s$-th stage of the encoders as an example and omit the superscript $s$ for simplicity. We change the channel numbers of both spatial features $S\in \mathbb{R}^{C\times H\times W}$ and temporal features $T\in \mathbb{R}^{C\times H\times W}$ to $C_m$ via linear transformation:
\begin{equation}
    T^1=\textit{Linear}(T),~~S^1=\textit{Linear}(S),
\end{equation}
where $S^1\in \mathbb{R}^{C_m\times H\times W}$ and $T^1\in \mathbb{R}^{C_m\times H\times W}$ are the visual inputs to the $1$-st LBDT layer.

For the linguistic inputs, we first enhance the words features $R\in \mathbb{R}^{N\times C_m}$ by the self-attention mechanism~\cite{vaswani2017attention} and denote the enhanced words features as $R^{\prime}\in \mathbb{R}^{N\times C_m}$, which can be formatted as follows:
\begin{equation}
\begin{aligned}
    &R^{Q/K/V}=\textit{Linear}(R+P_R),\\
    &R^{\prime}=\textit{Softmax}(\frac{R^Q(R^K)^T}{\sqrt{C_m}})R^V+R,
\end{aligned}
\end{equation}
where $P_R\in \mathbb{R}^{N\times C_m}$ is the sinusoids positional encoding \cite{vaswani2017attention}.

Our LBDT module follows the implementation practice of Transformer~\cite{vaswani2017attention} and revises it to a cross-modal version.
In each LBDT layer, we feed the enhanced language features $R^\prime$ and the outputs from the previous layer to it as inputs:
\begin{equation}
    S^{l+1},T^{l+1}=\textit{LBDT}(S^{l},T^{l},R^\prime), l=1,...,L-1.
\end{equation}

As the duplex transfer process happens in a symmetrical way, we take the \textit{temporal}$\rightarrow$\textit{language}$\rightarrow$\textit{spatial} transfer process in the $l$-th LBDT layer as an example (Figure~\ref{fig:lbdt}). For \textit{motion aggregation}, we first add the 2D sinusoids positional encoding $P_T\in \mathbb{R}^{C_m\times H\times W}$ to the temporal features $T^l$, and then reshape it to ${T^l}^\prime \in \mathbb{R}^{HW\times C_m}$. We obtain the attention map $A_{R\times T}\in \mathbb{R}^{N\times HW}$ by calculating the similarity between each word and each pixel:
\begin{equation}
    {T^l}^\prime=\textit{Reshape}(T^l+P_T),
\end{equation}
\begin{equation}
\begin{aligned}
    &R^Q = \textit{Linear}(R^\prime),~~T^K=\textit{Linear}({T^l}^\prime),\\
    &A_{R\times T} = \textit{Softmax}({\frac{R^Q{(T^K)}^T}{\sqrt{C_m}}}),
\end{aligned}
\end{equation}
where $A_{R\times T}=\{A_{R\times T}^i\}_{i=1}^N$ and $A_{R\times T}^i\in \mathbb{R}^{HW}$ is the attention map for the $i$-th word. We use $A_{R\times T}$ to adaptively aggregate the language-relevant motion information from the reshaped temporal features ${T^l}^\prime$, and then add it to the words features $R^\prime$ to obtain the \textit{language medium} $R_{T\rightarrow S}\in \mathbb{R}^{N\times C_m}$ with multimodal representations:

\begin{equation}
\begin{aligned}
    &T^V=\textit{Linear}({T^l}^\prime),\\
    &R_{T\rightarrow S} = A_{R\times T}T^V + R^\prime.
\end{aligned}
\end{equation}

For \textit{motion transfer}, we let the spatial features to adaptively select cross-modal motion information from the medium $R_{T\rightarrow S}$. Similarly, we first add the positional encoding $P_S$ to the spatial features $S^l$, and reshape it to ${S^l}^\prime \in \mathbb{R}^{HW\times C_m}$. Then we calculate the cross-attention map $A_{S\times R}^i \in \mathbb{R}^{N}$ between the $i$-th pixel of the spatial features and the medium $R_{T\rightarrow S}$, which measures the semantic relevance between these two features:
\begin{equation}
    {S^l}^\prime=\textit{Reshape}(S^l+P_S),
\end{equation}
\begin{equation}
\begin{aligned}
    R_{T\rightarrow S}^{K} &= \textit{Linear}(R_{T\rightarrow S}),~~S^Q = \textit{Linear}({S^l}^\prime),\\
    A_{S\times R}& = \textit{Softmax}(\frac{S^Q{(R_{T\rightarrow S}^K)}^T}{\sqrt{C_m}}).
\end{aligned}
\end{equation}
Afterwards, we transfer the language-relevant motion information to the spatial features with the cross-attention map $A_{S\times R}$:
\begin{equation}
\begin{aligned}
    &R_{T\rightarrow S}^{V} = \textit{Linear}(R_{T\rightarrow S}),\\
    &S^{l+1} = \textit{MLP}(A_{S\times R}R_{T\rightarrow S}^V) + S^l,
\end{aligned}
\end{equation}
where $\textit{MLP}$ denotes the multi-layer perception and $S^{l+1}$ is the output spatial features of the $l$-th LBDT layer.

We denote the outputs of the last LBDT layer $S^L$ and $T^L$ as the outputs of our LBDT module. Finally, we increase the channel numbers of $S^L$ and $T^L$ to $C$ and add them with original spatial and temporal features respectively to form a residual connection for easier optimization.

\subsection{Bilateral Channel Activation}
\label{sec:method:tfm}

\begin{figure}[t]
	\centering
		\includegraphics[width=\linewidth]{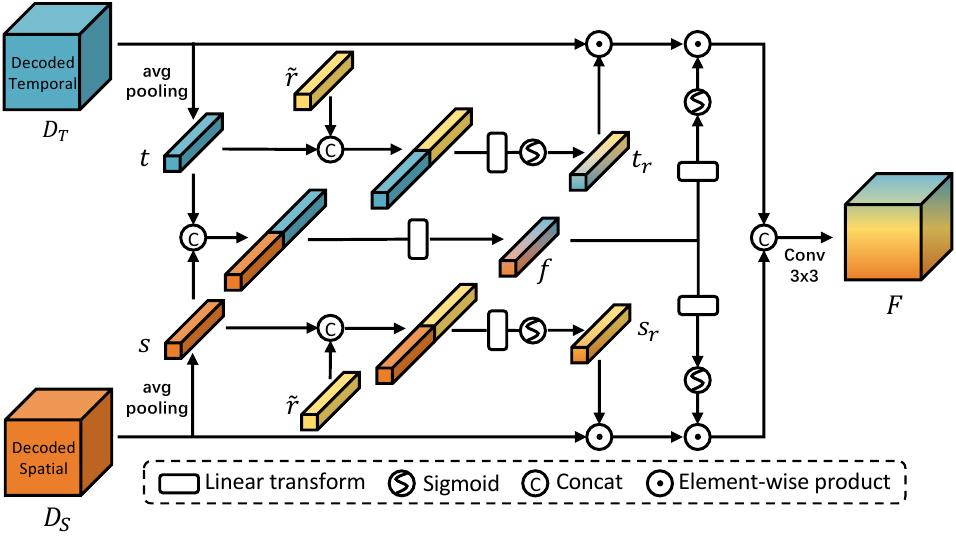}
	\caption{Illustration of BCA module. We filter out the language-irrelevant information with the language denoisers $t_r$ and $s_r$, and use $f_t$ and $f_s$ to highlight spatial-temporal consistent channels.}
	\label{fig:tfm}
	
\end{figure}

To obtain strong semantic representation and maintain local details of the frame simultaneously, we upsample the low-resolution spatial and temporal features in the last three stages $\{S_s\}_{s=3}^{5}$ and $\{T_s\}_{s=3}^{5}$ to the same size with the $2$-nd stage features $S_2$ and $T_2$.
The resulted features are denoted as $\{S_s^{up}\}_{s=3}^{5}\in \mathbb{R}^{C_d\times H_2\times W_2}$ and $\{T_s^{up}\}_{s=3}^{5}\in \mathbb{R}^{C_d\times H_2\times W_2}$ where $C_d$ is the channel number in the decoder.
Then, we add them with $S_2$ and $T_2$ to get the decoded features $D_S\in \mathbb{R}^{C_d\times H_2\times W_2}$ and $D_T\in \mathbb{R}^{C_d\times H_2\times W_2}$ for spatial and temporal decoders respectively.
Given $D_S$ and $D_T$, we also propose a Bilateral Channel Activation (BCA) module to adaptively filter out language-irrelevant information while highlighting consistent spatial-temporal features, which is illustrated in Figure~\ref{fig:tfm}.

Concretely, since $D_T$ and $D_S$ may contain the language-irrelevant motion and appearance information, we propose to exploit the sentence feature $\widetilde{r}=\sum_{n=1}^Nr_n \in \mathbb{R}^{C_r}$ as the denoiser to filter out the language-irrelevant information.
We first conduct average pooling on $D_S$ and $D_T$ to squeeze them into $s\in \mathbb{R}^{C_d}$ and $t\in \mathbb{R}^{C_d}$. Then, we obtain the language-specific spatial denoiser $s_r\in \mathbb{R}^{C_d}$ and temporal denoiser $t_r \in \mathbb{R}^{C_d}$ as follows:
\begin{equation}
    s_r=\sigma(\textit{Linear}([s;\widetilde{r}]),~~t_r=\sigma(\textit{Linear}([t;\widetilde{r}]),
\end{equation}
where $\sigma$ is sigmoid function and $[;]$ denotes concatenation.

Meanwhile, we also concatenate $s$ and $t$ on the channel dimension and apply linear transformation to obtain the spatial-temporal consistent feature $f\in \mathbb{R}^{C_d}$: 
\begin{equation}
    f=\phi(\textit{Linear}([t;s])),
\end{equation}
where $\phi(\cdot)$ is ReLU \cite{nair2010rectified} function. We transform $f$ to the channel activators $f_s\in \mathbb{R}^{C_d}$ and $f_t\in \mathbb{R}^{C_d}$ for $D_S$ and $D_T$ respectively using the sigmoid function $\sigma$:
\begin{equation}
    f_t=\sigma(\textit{Linear}(f)),~~f_s=\sigma(\textit{Linear}(f)).
\end{equation}

Next, the language-specific denoisers (\ie, $t_r$ and $s_r$) and spatial-temporal consistent activators (\ie, $f_t$ and $f_s$) are combined to process decoded spatial features $D_S$ and temporal feature $D_T$ before fusing them:
\begin{equation}
    D_T^\prime=f_t\odot t_r\odot D_T,~~D_S^\prime=f_s\odot t_s\odot D_S,
\end{equation}
where $\odot$ is the element-wise product with broadcasting operation.

\begin{table*}
    \centering
    \resizebox{\textwidth}{!}{
    \begin{tabular}{r||c|c|c|c|c|c|c|c|c}
    \hline\thickhline
    \rowcolor{mygray}
    & & \multicolumn{5}{c|}{Overlap} & AP & \multicolumn{2}{c}{IoU} \\
    \rowcolor{mygray}
    \multirow{-2}*{Method} & \multirow{-2}*{Pub.} & \multicolumn{1}{c}{P@0.5} & \multicolumn{1}{c}{P@0.6} & \multicolumn{1}{c}{P@0.7} & \multicolumn{1}{c}{P@0.8} & \multicolumn{1}{c|}{P@0.9} & 0.5:0.95 & \multicolumn{1}{c}{Overall} & Mean \\ \hline\hline
    Gavrilyuk \etal \cite{GavrilyukGLS18} & CVPR18 & 47.5 & 34.7 & 21.1 & 8.0 & 0.2 & 19.8 & 53.6 & 42.1 \\
    Gavrilyuk \etal $\dag$ \cite{GavrilyukGLS18} & CVPR18 & 50.0 & 37.6 & 23.1 & 9.4 & 0.4 & 21.5 & 55.1 & 42.6 \\
    ACGA \cite{WangDYT19} & ICCV19 & 55.7 & 45.9 & 31.9 & 16.0 & 2.0 & 27.4 & 60.1 & 49.0 \\
    VT-Capsule \cite{mcintosh2020visual} & CVPR20 & 52.6 & 45.0 & 34.5 & 20.7 & 3.6 & 30.3 & 56.8 & 46.0 \\
    CMDy \cite{WangDMY20} & AAAI20 & 60.7 & 52.5 & 40.5 & 23.5 & 4.5 & 33.3 & 62.3 & 53.1 \\
    PRPE \cite{NingXW020} & IJCAI20 & 63.4 & 57.9 & 48.3 & 32.2 & 8.3 & 38.8 & 66.1 & 52.9 \\
    CMSA \cite{ye2021referring} & TPAMI21 & 48.7 & 43.1 & 35.8 & 23.1 & 5.2 & - & 61.8 & 43.2 \\
    CSTM \cite{HuiH0DLWH021} & CVPR21 & 65.4 & 58.9 & 49.7 & 33.3 & 9.1 & 39.9 & 66.2 & 56.1 \\
    CMPC-V \cite{liu2021cross} & TPAMI21 & 65.5 & 59.2 & 50.6 & 34.2 & 9.8 & 40.4 & 65.3 & 57.3 \\\hline\hline
    \textbf{Ours (LBDT-1)} & - & \textbf{71.1 \color{sota_blue}{({+5.6})}}  & \textbf{66.1 \color{sota_blue}{({+6.9})}} & \textbf{57.8 \color{sota_blue}{({+7.2})}} & \textbf{41.6 \color{sota_blue}{({+7.4})}} &\textbf{12.0 \color{sota_blue}{({+2.2})}}  & \textbf{46.1 \color{sota_blue}{({+5.7})}} & \textbf{70.1 \color{sota_blue}{({+3.9})}} & \textbf{61.2 \color{sota_blue}{({+3.9})}} \\
     \textbf{Ours (LBDT-4)} & - & \textbf{73.0 \color{sota_blue}{({+7.5})}} & \textbf{67.4 \color{sota_blue}{({+8.2})}} & \textbf{59.0 \color{sota_blue}{({+8.4})}} & \textbf{42.1 \color{sota_blue}{({+7.9})}} & \textbf{13.2 \color{sota_blue}{({+3.4})}} & \textbf{47.2 \color{sota_blue}{({+6.8})}} & \textbf{70.4 \color{sota_blue}{({+4.2})}} &\textbf{62.1 \color{sota_blue}{({+4.8})}}  \\\hline
    \end{tabular}
    }
    \caption{Comparison with state-of-the-art methods on A2D Sentences testing set. Our method significantly outperforms previous methods relying on the 3D ConvNets for spatial-temporal interaction. ``$\dag$'' denotes utilizing additional optical flow inputs. ``LBDT-$x$'' indicates stacking $x$ LBDT layers in each LBDT module.}
    \label{exp:sota:a2d}
\end{table*}

\begin{table*}
    \centering
    \resizebox{\textwidth}{!}{
    \begin{tabular}{r||c|c|c|c|c|c|c|c|c}
    \hline\thickhline
    \rowcolor{mygray}
    & & \multicolumn{5}{c|}{Overlap} & AP & \multicolumn{2}{c}{IoU} \\
    \rowcolor{mygray}
    \multirow{-2}*{Method} & \multirow{-2}*{Pub.} & \multicolumn{1}{c}{P@0.5} & \multicolumn{1}{c}{P@0.6} & \multicolumn{1}{c}{P@0.7} & \multicolumn{1}{c}{P@0.8} & \multicolumn{1}{c|}{P@0.9} & 0.5:0.95 & \multicolumn{1}{c}{Overall} & Mean \\ \hline\hline
    Gavrilyuk \etal \cite{GavrilyukGLS18} & CVPR18 & 69.9 & 46.0 & 17.3 & 1.4 & 0.0 & 23.3 & 54.1 & 54.2 \\
    Gavrilyuk \etal $\ddagger$ \cite{GavrilyukGLS18} & CVPR18 & 71.2 & 51.8 & 26.4 & 3.0 & 0.0 & 26.7 & 55.5 & 57.0 \\
    ACGA \cite{WangDYT19} & ICCV19 & 75.6 & 56.4 & 28.7 & 3.4 & 0.0 & 28.9 & 57.6 & 58.4 \\
    VT-Capsule \cite{mcintosh2020visual} & CVPR20 & 67.7 & 51.3 & 28.3 & 5.1 & 0.0 & 26.1 & 53.5 & 55.0 \\
    CMDy \cite{WangDMY20} & AAAI20 & 74.2 & 58.7 & 31.6 & 4.7 & 0.0 & 30.1 & 55.4 & 57.6 \\
    PRPE \cite{NingXW020} & IJCAI20 & 69.0 & 57.2 & 31.9 & 6.0 & 0.1 & 29.4 & - & - \\
    CMSA \cite{ye2021referring} & TPAMI21 & 76.4 & 62.5 & 38.9 & 9.0 & 0.1 & - & 62.8 & 58.1 \\
    CSTM \cite{HuiH0DLWH021} & CVPR21 & 78.3 & 63.9 & 37.8 & 7.6 & 0.0 & 33.5 & 59.8 & 60.4 \\ 
    CMPC-V \cite{liu2021cross} & TPAMI21 & 81.3 & 65.7 & 37.1 & 7.0 & 0.0 & 34.2 & 61.6 & 61.7 \\\hline\hline
    \textbf{Ours (LBDT-1)} & - & \textbf{86.4 \color{sota_blue}{({+5.1})}}  & \textbf{75.1 \color{sota_blue}{({+9.4})}} & \textbf{50.7 \color{sota_blue}{({+11.8})}} & \textbf{11.6 \color{sota_blue}{({+2.6})}} & 0.1 & \textbf{40.3 \color{sota_blue}{({+6.1})}} & \textbf{64.6 \color{sota_blue}{({+1.8})}} & \textbf{65.2 \color{sota_blue}{({+3.5})}} \\
     \textbf{Ours (LBDT-4)} & - & \textbf{86.4 \color{sota_blue}{({+5.1})}} & \textbf{74.4 \color{sota_blue}{({+8.7})}} & \textbf{53.3 \color{sota_blue}{({+14.4})}} & \textbf{13.2 \color{sota_blue}{({+4.2})}} & 0.0 & \textbf{41.1 \color{sota_blue}{({+6.9})}} & \textbf{64.5 \color{sota_blue}{({+1.7})}} &\textbf{65.8 \color{sota_blue}{({+4.1})}}  \\\hline
    \end{tabular}
    }
    \caption{Comparison with state-of-the-art methods on J-HMDB Sentences dataset using the best model trained on A2D Sentences. Our method shows notable generalization ability. $\ddagger$ denotes training more layers of I3D backbone on A2D Sentences.}
    \label{exp:sota:jhmdb}
\end{table*}

Finally, we concatenate the refined spatial feature $D_S^\prime$ and temporal feature $D_T^\prime$ together and use the 3$\times$3 convolution to obtain the fused feature $F\in \mathbb{R}^{C_d\times H_2\times W_2}$. We further apply convolutions and sigmoid function on $F$ to obtain the logit map and up-sample it to the same spatial size with inputs as the prediction $P\in \mathbb{R}^{1\times H_0\times W_0}$.

\section{Experiments}
\label{sec:exp}

\subsection{Datasets and Evaluation Criteria}
\label{sec:exp:criteria}

We evaluate the performance of our method on four popular referring video object segmentation benchmarks: A2D Sentences~\cite{GavrilyukGLS18}, J-HMDB Sentences~\cite{GavrilyukGLS18}, Refer-YouTube-VOS~\cite{seo2020urvos}, and Refer-DAVIS$_{\text{17}}$~\cite{khoreva2018video}.
For A2D Sentences and J-HMDB Sentences, we use IoU and Precision@X (P@X) as the evaluation criteria following~\cite{HuiH0DLWH021,NingXW020,WangDMY20}.
Specifically, the overall IoU is the ratio of the total intersection area divided by the total union area over all testing samples, and the mean IoU is the averaged IoU of all testing samples. P@X measures the percentage of test samples whose IoU is higher than a predefined threshold X, where $\text{X}\in \{0.5, 0.6, 0.7, 0.8, 0.9\}$. We also compute the average precision (AP) over the interval of $[0.50:0.05:0.95]$. For Refer-YouTube-VOS and Refer-DAVIS$_{\text{17}}$, we use region similarity ($\mathcal{J}$) and contour accuracy ($\mathcal{F}$) following~\cite{seo2020urvos}.

\begin{table}
    \centering
    \resizebox{1\linewidth}{!}{
    \begin{tabular}{r||c|c|c|c}
    \hline\thickhline
    \rowcolor{mygray}
    Method & Pub. & $\mathcal{J}$ & $\mathcal{F}$ & $\mathcal{J}\&\mathcal{F}$ \\ \hline\hline
    URVOS $\dag$ \cite{seo2020urvos} & ECCV20 & 41.34 & - & - \\ 
    URVOS \cite{seo2020urvos} & ECCV20 & 45.27 & 49.19 & 47.23 \\
    CMPC-V \cite{liu2021cross} & TPAMI21 & 45.64 & 49.32 & 47.48 \\ \hline\hline
    \textbf{Ours (LBDT-4)} & - & \textbf{48.18 \color{sota_blue}{({+2.54})}} & \textbf{50.57 \color{sota_blue}{({+1.25})}} & \textbf{49.38 \color{sota_blue}{({+1.90})}} \\\hline
    \end{tabular}
    }
    \caption{Comparison with state-of-the-art methods on the Refer-Youtube-VOS validation set. $\dag$ indicates removing multiple iterations of the second stage inference step.}
    \label{exp:sota:urvos}
\end{table}

\begin{table}
    \centering
    \resizebox{1\linewidth}{!}{
    \begin{tabular}{r||c|c|c|c}
    \hline\thickhline
    \rowcolor{mygray}
    & & & \multicolumn{2}{c}{$\mathcal{J}\&\mathcal{F}$}\\
    \rowcolor{mygray}
    \multirow{-2}*{Method} & \multirow{-2}*{Pub.} & \multirow{-2}*{Pretrained}  & \multicolumn{1}{c}{1st video} & \multicolumn{1}{c}{full video} \\ \hline\hline
    Khoreva~\etal~\cite{khoreva2018video} & ACCV18 & RefCOCO \cite{nagaraja2016modeling} & 39.30 & 37.10 \\ 
    URVOS \cite{seo2020urvos} & ECCV20 & RefCOCO \cite{nagaraja2016modeling} & 44.10 & - \\
 
    URVOS \cite{seo2020urvos} & ECCV20 & Refer-Youtube-VOS \cite{seo2020urvos} & 51.63 & - \\ \hline\hline
    \textbf{Ours (LBDT-4)} & - & Refer-Youtube-VOS \cite{seo2020urvos} & \textbf{54.08 \color{sota_blue}{({+2.45})}} & \textbf{54.52 \color{sota_blue}{({+17.42})}}  \\\hline
    \end{tabular}
    }
    \caption{Comparison with state-of-the-art methods on the Refer-DAVIS$_{17}$ validation set.}
    \label{exp:sota:davis}
\end{table}

\subsection{Implementation Details}
\label{sec:exp:details}

\begin{table*}
    \centering
    {
    \tablestyle{6pt}{1.15}\begin{tabular}{c|c|c|c||c|c|c|c|c|c|c|c}
    \hline\thickhline
    \rowcolor{mygray}
    \multicolumn{2}{c|}{LBDT} & \multicolumn{2}{c||}{BCA} & \multicolumn{5}{c|}{Overlap} & AP & \multicolumn{2}{c}{IoU} \\
    \rowcolor{mygray}
    \multicolumn{1}{c}{S$\rightarrow$L$\rightarrow$T} & \multicolumn{1}{c|}{T$\rightarrow$L$\rightarrow$S} & \multicolumn{1}{c}{LD} & \multicolumn{1}{c||}{STC} & \multicolumn{1}{c}{P@0.5} & \multicolumn{1}{c}{P@0.6} & \multicolumn{1}{c}{P@0.7} & \multicolumn{1}{c}{P@0.8} & \multicolumn{1}{c|}{P@0.9} & 0.5:0.95 & \multicolumn{1}{c}{Overall} & Mean \\ \hline\hline
     & & & & 60.5 & 54.9 & 47.8 & 35.0 & 11.0 & 38.8 & 61.6 & 54.5 \\ \hline\hline
      \checkmark & & &  & 64.5 & 58.8 & 49.8 & 35.9 & 11.2 & 40.6 & 67.3 & 56.1 \\
    & \checkmark & & & 68.0 & 63.2 & 54.8 & 38.4 & 11.9 & 43.7 & 68.6 & 58.3 \\ 
  \checkmark & \checkmark & & & 70.0 & 64.1 & 55.7 & 39.3 & 11.5 & 44.5 & 69.3 & 59.8  \\ \hline\hline
  \checkmark & \checkmark &  & \checkmark & 70.3 & 65.3 & 56.7 & 40.4 & 12.2 & 45.3 & 69.5 & 60.4 \\
  \checkmark & \checkmark & \checkmark &  & 70.2 & 64.7 & 56.5 & 39.9 & \textbf{12.3} & 45.0 & 69.9 & 59.9 \\
    \checkmark & \checkmark & \checkmark & \checkmark & \textbf{71.1}  & \textbf{66.1} & \textbf{57.8} & \textbf{41.6} & 12.0 & \textbf{46.1} & \textbf{70.1} & \textbf{61.2} \\ \hline
    \end{tabular}
    }
    \caption{Verification of the effectiveness of our proposed LBDT module and BCA module. ``S$\rightarrow$L$\rightarrow$T'' and ``T$\rightarrow$L$\rightarrow$S'' denote \textit{spatial}$\rightarrow$\textit{language}$\rightarrow$\textit{temporal} transfer and \textit{temporal}$\rightarrow$\textit{language}$\rightarrow$\textit{spatial} transfer respectively. ``LD'' and ``STC'' denote \textit{language denoiser} and \textit{spatial-temporal consistent activator}.}
    \label{tab:ablation:parts}
\end{table*}

We use ResNet-50~\cite{He2016CVPR} pretrained on ImageNet~\cite{krizhevsky2012imagenet} dataset as our spatial and temporal encoders. For linguistic inputs, we adopt LSTM~\cite{hochreiter1997long} to extract language features from GloVe word embeddings~\cite{pennington2014glove} pretrained on Common Crawl with 840B tokens. The maximum length of the input sentence is 25. We set the frame interval $\delta=6$ for calculating the frame difference unless otherwise stated. The input frames are resized to $320\times 320$. Adam \cite{kingma2014adam} is utilized as the optimizer. We train the whole network in an end-to-end way with batch size 8 and learning rate $1e^{-4}$ for 15 epochs on NVIDIA Tesla V100 GPUs, which is supervised by cross-entropy loss and dice loss \cite{milletari2016v}. The learning rate is divided by 2 for every 2 epochs started from the $10$-th epoch. As for Refer-DAVIS$_{\text{17}}$, we finetune the best model trained on Refer-Youtube-VOS for 1 epoch with a learning rate $1e^{-5}$.

\subsection{Comparison with State-of-the-Art Methods}

We compare our method with previous state-of-the-art methods on four popular benchmarks mentioned before.
As shown in Table~\ref{exp:sota:a2d}, our method outperforms previous works by large margins on the A2D Sentences test set~\cite{GavrilyukGLS18}. Compared with CSTM~\cite{HuiH0DLWH021}, our LBDT-1 model achieves 5.7\%, 3.9\%, and 3.9\% absolute improvements on AP, Overall IoU, and Mean IoU respectively, indicting that using language as the medium to conduct explicit spatial-temporal interaction in the encoding phase is superior to existing methods using 3D ConvNets and implicit interaction in the decoding phase. By stacking LBDT layers, spatial and temporal features can be iteratively optimized, and the best performance is obtained using the LBDT-4 model with 4 layers.

We further verify the generalization ability of our method on J-HMDB Sentences dataset~\cite{GavrilyukGLS18}.
Following prior works~\cite{HuiH0DLWH021,NingXW020,WangDMY20}, we use the best model trained on A2D Sentences to directly evaluate all the samples in J-HMDB Sentences without finetuning.
As shown in Table~\ref{exp:sota:jhmdb}, our method accomplishes significant performance gains over previous state-of-the-arts, indicating that our method can obtain more robust multi-modal representations and generalize the learned knowledge to unseen datasets.

We also conduct experiments on the newly proposed Refer-YouTube-VOS benchmark~\cite{seo2020urvos} with richer object categories and denser annotated frames.
As shown in Table~\ref{exp:sota:urvos}, our method outperforms URVOS~\cite{seo2020urvos} and CMPC-V \cite{liu2021cross} by 2.15\% and 1.90\% on the $\mathcal{J}\&\mathcal{F}$ metrics respectively, demonstrating that our approach can perform well even in complex scenarios.  Moreover, following URVOS \cite{seo2020urvos}, we use the best model trained on the Refer-YouTube-VOS and finetune it on the Refer-DAVIS$_{17}$ dataset~\cite{khoreva2018video}, where we also achieve the best performance as shown in Table~\ref{exp:sota:davis}.

\subsection{Ablation Studies}

We conduct ablation studies on the A2D Sentences dataset to evaluate the different designs of our model.
All experiments are based on our LBDT-1 model.

\textbf{Component Analysis.} We summarize the ablation results of our proposed components in Table~\ref{tab:ablation:parts}.
The $1$-st row is the baseline method, where we first fuse the language features with the visual features and then conduct the duplex spatial-temporal interaction with the cross-attention mechanism~\cite{vaswani2017attention} directly between spatial and temporal features without the bridging of language, whose computational complexity is $\mathcal{O}((HW)^2C)$.
As shown in the $2$-nd and 3rd rows, both \textit{motion transfer} and \textit{appearance transfer} can bring notable improvements, validating the effectiveness of the language-bridged duplex strategy.
Moreover, the complexity of our LBDT module is $\mathcal{O}(HWNC)$, which is more lightweight as $N\ll HW$.
As for the BCA module, we conduct ablation experiments on two key components: \textit{language denoiser} and \textit{spatial-temporal consistency activator}.
It shows the decoded spatial and temporal features can benefit each other by activating the spatial-temporal consistent channels (the $5$-th row), and reducing the language-irrelevant motion and appearance information can improve the vanilla feature fusion in the decoder (the $6$-th row).

\begin{table}[!htbp]
    \centering
    \resizebox{\linewidth}{!}{
    \begin{tabular}{r||c|c|c|c|c|c|c}
    \hline\thickhline
    \rowcolor{mygray}
    & \multicolumn{7}{c}{Interval}\\
    \rowcolor{mygray}
    \multirow{-2}*{Metrics} & \multicolumn{1}{c}{1} & \multicolumn{1}{c}{2} & \multicolumn{1}{c}{3} & \multicolumn{1}{c}{4} & \multicolumn{1}{c}{5} & \multicolumn{1}{c}{6} & \multicolumn{1}{c}{7} \\ \hline\hline
    AP & 45.0 & 45.2 & 45.1 & 45.6 & 45.0 & \textbf{46.1} & 45.2 \\
    Mean & 68.9 & 68.7 & 69.4 & 69.3 & 69.0 & \textbf{70.1} & 69.0\\
    Overall & 60.0 & 60.5 & 60.1 & 60.8 & 60.0 & \textbf{61.2} & 60.8 \\ \hline
    \end{tabular}
    }
    \caption{Interval for calculating the frame difference. ``Mean'' and ``Overall'' are Mean IoU and Overall IoU respectively.}
    
    \label{tab:ablation:interval}
\end{table}

\textbf{Interval for Calculating the Frame Difference.} We demonstrate the influence of the interval value $\delta$ for calculating the frame difference in Table~\ref{tab:ablation:interval}. We found that the best performance is achieved when the interval is 6, which achieves a balance of modeling short and long actions.

\begin{table}[!htbp]
    \centering
    {
    \tablestyle{8pt}{1}\begin{tabular}{c|c|c|c||c|c|c}
    \hline\thickhline
    \rowcolor{mygray}
    \multicolumn{4}{c||}{Stages} & AP & \multicolumn{2}{c}{IoU}\\
    \rowcolor{mygray}
    \multicolumn{1}{c}{2} & \multicolumn{1}{c}{3} & \multicolumn{1}{c}{4} & \multicolumn{1}{c||}{5} & 0.5:0.95 & \multicolumn{1}{c}{Overall} & \multicolumn{1}{c}{Mean} \\ \hline\hline
    & & & \checkmark & 42.3 & 67.3 & 57.8 \\
    & & \checkmark & \checkmark & \textbf{46.1} & \textbf{70.1} & \textbf{61.2} \\
    & \checkmark & \checkmark & \checkmark & 45.3 & 68.2 & 60.8  \\
    \checkmark & \checkmark & \checkmark & \checkmark & 38.7 & 63.7 & 55.9 \\ \hline
    \end{tabular}
    }
    \caption{Inserting stages of LBDT module.}
    
    \label{tab:ablation:stages}
\end{table}

\textbf{Inserting Stages of LBDT Module.} We evaluate different inserting positions of LBDT module and summarize the results in Table~\ref{tab:ablation:stages}. Inserting LBDT into the $4$-th and $5$-th stages of our spatial and temporal encoders can bring significant improvements, but the performance decreases as we insert it into the earlier stages (\ie, $2$-nd and $3$-rd stages). It indicates that the language-bridged spatial-temporal interaction is more suitable for transferring the high-level semantic information.

\begin{table}[!htbp]
    \centering
    {
    \tablestyle{7pt}{1.14}\begin{tabular}{r||c|c|c|c}
    \hline\thickhline
    \rowcolor{mygray}
    Method & Input Size & GFLOPs & FPS & AP \\ \hline\hline
    ACGA~\cite{WangDYT19} & 16$\times$512$\times$512 & 630.83 & 9.5 & 27.4 \\
    CSTM~$\dag$~\cite{HuiH0DLWH021} & 8$\times$320$\times$320 & 213.06 & 11.4 & 39.9 \\ \hline\hline
    Ours (LBDT-1) & 2$\times$320$\times$320 & \textbf{32.51} & \textbf{19.2} & 46.1\\ 
    Ours (LBDT-4) & 2$\times$320$\times$320 & 38.03 & 12.5  & \textbf{47.2} \\ \hline
    \end{tabular}
    }
    \caption{Computational overhead. The RGB input size is \textit{frames} $\times$ \textit{height} $\times$ \textit{width} (channels are omited). $\dag$ denotes the inference code is obtained by contacting the authors.}
    \label{tab:ablation:flops}
\end{table}

\subsection{Computational Overhead}
We compare the computational overhead of our method and previous ones in Table~\ref{tab:ablation:flops}. Without the dependency on 3D ConvNet, our model outperforms existing methods by significant margins while consuming around 7$\times$ less GFLOPs and a much smaller input size. Moreover, we evaluate the FPS of these methods on a single NVIDIA 1080Ti GPU. It shows that our method is more efficient, which increases the possibility of practical applications for RVOS.

\subsection{Qualitative Analysis}
\label{exp:visualize}

\begin{figure}[t]
	\centering
		\includegraphics[width=\linewidth]{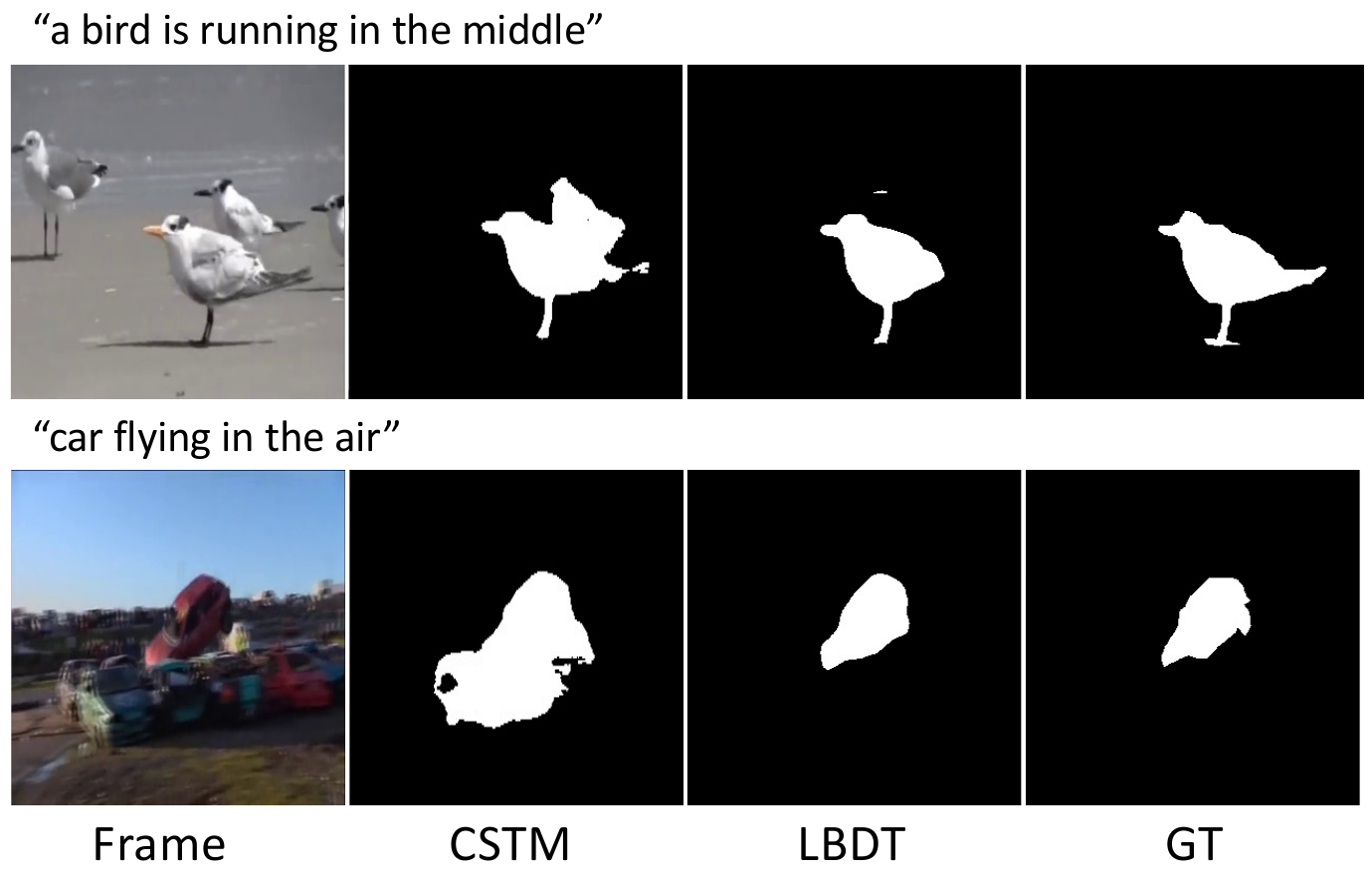}
	\caption{Visualization of the predicted masks of ours and CSTM \cite{HuiH0DLWH021} in the complex scenes.}
    
	\label{fig:sota}
	
\end{figure}

\begin{figure}[t]
	\centering
		\includegraphics[width=\linewidth]{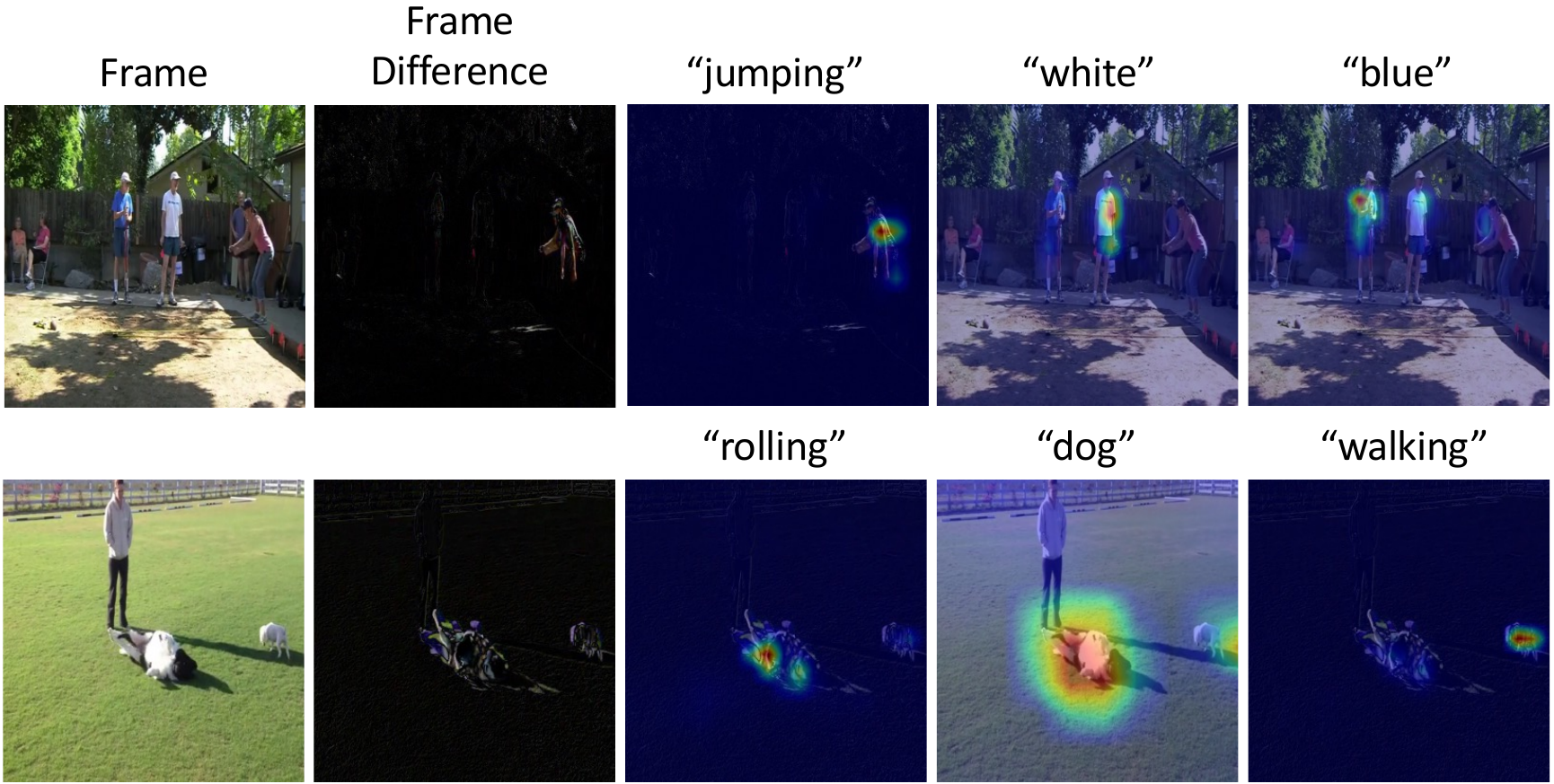}
	\caption{Visualization of attended regions of the referring words.}

	\label{fig:attn}
	
\end{figure}

Figure~\ref{fig:sota} presents the predictions of our method and CSTM \cite{HuiH0DLWH021} in the complex scenes. As the spatial encoder in CSTM lacks motion information, it tends to generate masks on false objects ($2$-nd column). By explicitly conducting spatial-temporal interaction in the encoding phase with language as the bridge, our methods can obtain the accurate masks of the referred objects ($3$-rd column).
We further visualize the attended regions of the referring words in our LBDT module in Figure~\ref{fig:attn}. Taking the $1$-st row as an example, the motion-related word ``jumping'' attends to the region of the jumping girl, and the appearance-related word ``white'' and ``blue'' has the highest responses on the two people in the corresponding colors. 

\section{Conclusion and Discussion}
\label{sec:conclu}

In this paper, we reconsider the way of spatial-temporal interaction for RVOS and propose a Language-Bridged Duplex Transfer (LBDT) module to explicitly conduct spatial-temporal interaction in the encoding phase with language as the medium for transferring the language-relevant information. A Bilateral Channel Activation (BCA) module is also introduced in the decoding phase to denoise and activate the spatial-temporal consistent features via channel activation. Experiments show that our methods outperform previous methods by large margins on four popular benchmarks with much less computational overhead.

\paragraph{Limitation.} The limitation of our paper is that static language descriptions may not always match the dynamic objects whose location and pose are various in continuous frames. In the future, we hope to address the mentioned mismatch problem by exploring the temporal coherence between the masks in different video frames, which is complementary to our focus
in this paper.

\paragraph{Acknowledgments} This research is partly supported by National Natural Science Foundation of China (Grant 62122010, 61876177), Fundamental Research Funds for the Central Universities, and Key R\&D Program of Zhejiang Province (2022C01082). 

{\small
\bibliographystyle{ieee_fullname}
\bibliography{egbib}
}

\end{document}